\documentclass[prodmode,acmtecs]{acmsmall} 

\usepackage[ruled]{algorithm2e}

\usepackage{CJK}
\usepackage{graphicx}
\usepackage{amsmath}
\usepackage{color}
\usepackage{sistyle}
\SIthousandsep{,}
\usepackage{tikz}
\usepackage{pgfplots}
\definecolor{gallery}{RGB}{240,240,240}
\definecolor{mercury}{RGB}{230,230,230}
\definecolor{tuatara}{RGB}{67, 67, 67}
\definecolor{shakespeare}{RGB}{85, 154, 193}
\definecolor{sun_shade}{RGB}{255, 144, 68}
\definecolor{free_speech_aquamarine}{RGB}{0, 156, 114}
\definecolor{flamingo}{RGB}{237, 88, 85}

\SetAlFnt{\small}
\SetAlCapFnt{\small}
\SetAlCapNameFnt{\small}
\SetAlCapHSkip{0pt}
\IncMargin{-\parindent}

\acmVolume{9}
\acmNumber{4}
\acmArticle{39}
\acmYear{2010}
\acmMonth{3}

\doi{0000001.0000001}

\issn{1234-56789}

\begin{document}

\markboth{HN. Zhang et al.}{Neural or Statistical: An Empirical Study on Language Models for Chinese Input Recommendation on Mobile}
\title{Neural or Statistical: An Empirical Study on Language Models for Chinese Input Recommendation on Mobile}
\author{HAINAN ZHANG, YANYAN LAN, JUN XU, JIAFENG GUO, XUEQI CHENG
\affil{Institute of Computing Technology, Chinese Academy of Sciences}
}

\begin{abstract}
Chinese input recommendation plays an important role in alleviating human cost in typing Chinese words, especially in the scenario of mobile applications.
The fundamental problem is to predict the conditional probability of the next word given the sequence of previous words.
Therefore, statistical language models, i.e.~n-grams based models, have been extensively used on this task in real application. However, the characteristics of extremely different typing behaviors usually lead to serious sparsity problem, even n-gram with smoothing will fail. A reasonable approach to tackle this problem is to use the recently proposed neural models, such as probabilistic neural language model, recurrent neural network and word2vec. They can leverage more semantically similar words for estimating the probability. However, there is no conclusion on which approach of the two will work better in real application. In this paper, we conduct an extensive empirical study to show the differences between statistical and neural language models. The experimental results show that the two different approach have individual advantages, and a hybrid approach will bring a significant improvement.
\end{abstract}

%
%

\ccsdesc[500]{Computer systems organization~Embedded systems}
\ccsdesc[300]{Computer systems organization~Redundancy}
\ccsdesc{Computer systems organization~Robotics}
\ccsdesc[100]{Networks~Network reliability}

%
%


\keywords{Neural network, deep learning, language model, machine learning, sequential prediction}

\acmformat{Hainan Zhang, Yanyan Lan, Jun Xu, Jiafeng Guo and Xueqi Cheng, 2016. Neural or Statistical: An Empirical Study on Language Models for Chinese Input Recommendation on Mobile.}
\begin{bottomstuff}
This work is supported by the National Science Foundation, under
grant CNS-0435060, grant CCR-0325197 and grant EN-CS-0329609.
Author's addresses: Web Data Science and Engineering Lab, Institute of Computing Technology, Chinese Academy of Sciences.
\end{bottomstuff}
\maketitle
\section{Introduction}
Chinese input recommendation is a useful technology in reducing users' human cost in typing Chinese words, especially when users are using some mobile applications, since the typing costs are heavier than those on the computer. Typically, the recommender system will recommend some possible next words based on a user' typing history, and the user can directly select the word which he or she would like to input rather than directly typing it. Therefore, the typing costs can be largely saved. This task can usually be formulated as follows: given a user's typing history, i.e.~a word sequence $w_1,\cdots,w_{t-1}$ for example, the recommendation engine will provide a recommendation list according to the probability of generating the next word $w_t$. Therefore, the fundamental problem is to estimate the conditional probability of the next word $w_t$ given the word sequence $w_1,\cdots,w_{t-1}$, i.e.~$P(w_t|w_1,\cdots,w_{t-1})$. This is exactly the target of language models \cite{GAO:INPUTTYPING,SI:machinetranslation,SP:JF}.

Traditional n-gram based statistical language models can be a natural choice to this task, based on the assumption that a user would like to input the same sequential patterns which other users have been input. Benefiting from the simplicity and stability, n-gram based statistical language models have been widely used in the real application of Chinese input recommendation. However, it cannot fully solve this problem since the task of Chinese input recommendation has some intrinsic characteristics. (1) The representation of Chinese input can be quite diverse, even though users refer to the same meaning. (2) Users' typing behaviors are quite different. Some users input Chinese sentences character by character, while others prefer the way of word by word, or even directly typing the Pinyin of the whole sentence. These characteristics pose great challenges of sparsity to traditional n-gram based models, even smoothed n-gram \cite{CS:SMOOTHMETHODS,KN:KNSMOOTH} will fail in recommending the useful next words.

Recently in academic community, a new approach named neural language models has been proposed to further taking into account the `similarity' between words. Examples include neural language model (NLM for short) \cite{Bengio:FFNNLM}, word2vec \cite{MK:EMBEDDING,Mikolov:word2vec2}, recurrent neural network \cite{MK:RNN} (RNN for short) and long short term memory \cite{H:LSTM} (LSTM for short). Specifically, these models represent each word as a dense vector. Therefore, different words can be connected by their semantic representations to alleviate the sparsity problem. So far however, there has been no conclusion on which one of the two different approach will perform better and should be used in real application, to the best of our knowledge.

In this paper, we conduct an extensive empirical study on real data of Chinese input recommendation to compare the two different approaches.
Specifically, we collect a large scale data set, named CIR from a commercial company focusing on Chinese input recommendation to facilitate our study. The data set contains 20,000,000 word sequences, and each sequence is composed of 4.9 words on average. In our study, we don't use any pre-process of Chinese word segmentation since we do not want to be biased by any artificial segmentation in the real application.

The experimental results show that when using a single language model for the task of Chinese input recommendation, the statistical language model can give the best results, the neural probabilistic language model is a little worse, while the other neural language models perform the worst. This result tells us that the exact matching approach (n-gram based models) can find accurate results, while semantic matching approach (neural language models) can give further candidates but also introduce noises.
We also find that the overlap between the results given by two different approaches are relatively small. Therefore, it motivates us to use a hybrid model to combine both approaches, and thus better recommendation results can be obtained than using the single model.
Specifically, the combination of NLM with n-gram model is better than that of other neural models and n-grams.
We also find that  word2vec is different with NLM and RNN due to the negative sampling, and the combination of n-gram, word2vec and NLM can obtains further improvement.

The rest of the paper is organized as follows. Section $2$ discusses the background of Chinese input recommendation task, the statistical language model and the evaluation measures. Section $3$ describes the neural language models, including NLM, word2vec, RNN, and LSTM.
Section $4$ shows our experimental settings. Section $5$ shows the experimental results and discussions. The conclusion is made in Section $6$.
\section{Backgrounds on Chinese Input Recommendation}\label{sec:backgrounds}
In this section, we will introduce some backgrounds on Chinese input recommendation, including the task description, state-of-the-art algorithms and evaluation measures.
\subsection{Task Description}
\begin{figure}
\centering
\includegraphics[height=2.1in, width=2.1in]{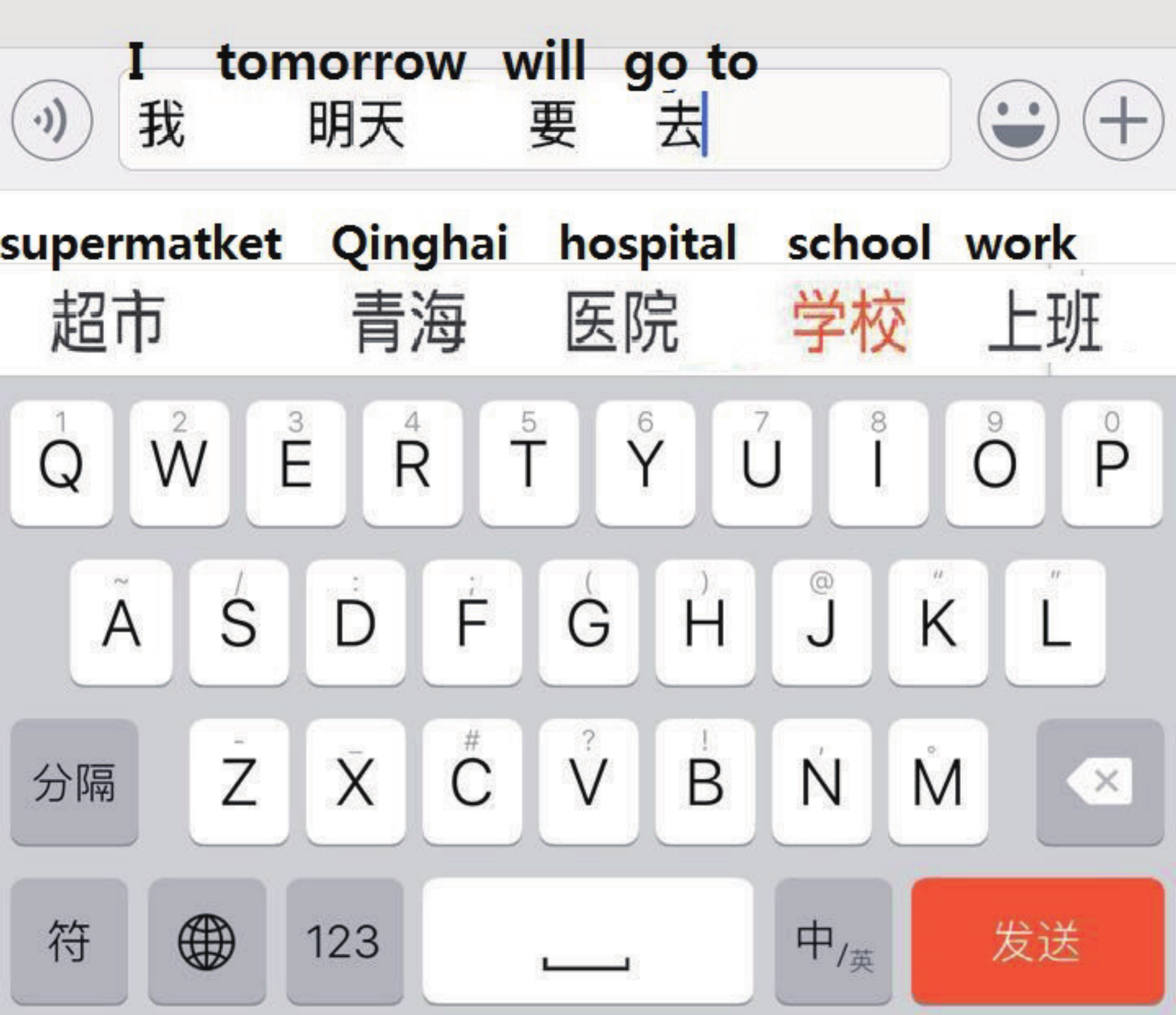}
\caption{\label{fig:taskmobile} An example of Chinese input recommendation on the mobile phone.}
\end{figure}
\begin{CJK*}{UTF8}{gbsn}

Inputing Chinese words is difficult on the mobile device. Taking the most popular Pinyin\cite{CS:PINYIN} typing methods for example, people first need to type the right Pinyin (composed of different English characters), and then choose which words he/she wants to input since different words can share a same pinyin representation. For example, if a user want to input a Chinese word `推导(deduce)', he needs to first type a Pinyin `tuidao', and then choose the word `推导(deduce)' from `推倒(push over)' `退到(back to)' and `推导(deduce)'. Or if the user is in favor of typing the Chinese character one by one, he may first type the Pinyin `tui' and choose the character ` 推(push)' from `推(push)' `退(back)' `忒(very)', and then type the Pinyin `dao' and select the character `导(deduce)' from `到(to)' `导(deduce)' `道(way)' and `倒(over)', etc. We can see that this is more difficult than English words input task on mobile. That's why Chinese input recommendation becomes an important tool for mobile devices. Based on users' typing history, the recommender system can give a ranking list of possible words user want to type in the next step. With this ranking list, users can directly input the words by clicking rather than typing if the words is recommended on top of the ranking list. By this way, the typing efforts can be largely reduced and the typing error can be avoided at the same time. Therefore, we can see that an effective Chinese input recommendation algorithm is critical on mobile, since the screen of mobile is usually very small and only few top recommended words can be displayed on the screen.

Figure~\ref{fig:taskmobile} gives an illustration of the real system of Chinese input recommendation. In the example, we are given a user's typing history, i.e.~a sequence of words `我(I)-明天(tomorrow)-要(will)-去(go to)'. The recommender system gives a recommendation list containing five words: `超市(supermarket) 青海(Qinghai) 医院(hospital) 上班(work) 学校(school)'. If the user's next word exactly lies in the recommendation list, he/she can directly click the word (e.g.~`学校'(school)), rather than typing the exact words. Therefore, we can see that users' cost is largely reduced. This is extremely useful in the scenario of mobile.
\end{CJK*}

Therefore, the task of Chinese input recommendation can be formulated as a language modeling problem. Given a user's typing history, i.e. a word sequence $w_1,\cdots,w_{t-1}$, a recommendation list can be given out by estimating the conditional probability of each next word $P(w_t|w_1,\cdots,w_{t-1})$. The goal is to rank the required words on top of the ranking lists, i.e.~$P(w_t^*|w_1,\cdots,w_{t-1})=\max_{w_t} P(w_t|w_1,\cdots,w_{t-1})$, where $w_t*$ is the exact next words the user want to input.

\subsection{Statistical Language Models}
N-gram based models\cite{KS:NGRAM} are the most popular and successful statistical language models, which has also been widely used in the real application of Chinese input recommendation due to its simplicity in implementation and explanation. The basic assumption is that the probability of a word only depends on the preceding $n-1$ words. Therefore, the probability generated for a specific word sequence $w_1,\cdots,w_t$ is calculated as:
$$P(w_1,\cdots,w_t)\approx \prod_{i=1}^t P(w_i|w_{i-(n-1)},\cdots,w_{i-1}).$$
The conditional probability can then be calculated from the frequency counts, based on the maximum likelihood estimation.
\begin{displaymath}
P(w_i|w_{i-{n-1}},{\cdots},w_{i-1}){=}\frac{count(w_{i-{n-1}},\cdots,w_i)}{count(w_{i-{n-1}},\cdots,w_{i-1})}.
\end{displaymath}
Typically, however, the n-gram based probabilities are not derived directly from the frequency counts, because models derived this way have severe sparsity problems. That is to say, some n-grams may have not explicitly been seen before. Instead, some form of smoothing strategy is proposed to tackle this problem, by assigning some of the total probability mass to unseen words or n-grams. In this paper, we use a popular smoothing method, namely the interpolated Kneser-Ney smoothing\cite{Bengio:FFNNLM,KN:KNSMOOTH,KN:KNSMOOOTH2,Stolcke:srilm2002,Stolcke:srilm2011}, for comparison.  Let us denote the context as $c=w_{i-{n-1}},\cdots,w_{i-1}$, the interpolated Kneser-Ney estimates the conditional probability by discounting the true $count(c,w_{i})$ with a fixed amount $d(c)$, depending on the length $(c,w)$ if $count(c,w)>0$ (otherwise the count remains 0). Furthermore, it interpolates the estimated probability of word $w_i$ with lower order m-gram probabilities. The mathematical description is given as follows:
\begin{displaymath}
P(w_i|c){=}\frac{max(0,count(c,w_i)-d(c))}{count(c)}+\frac{d(c)t(c)}{count(c)}P'(w_i|c)
\end{displaymath}
where $t(c)=\#\{w'|count(c,w')>0\}$ is the number of distinct words $w'$ following context $c$ in training data, and the $P'(w_i|c)$ are the lower order m-gram probabilities.

\subsection{Evaluation Measures}
Typical measures for recommendation task is Precision, Recall, and F1 score\cite{CHEN:MAP,V:MAP2}. We give their formal definitions as follows.

Assume that for a given context $w_1,\cdots,w_{t-1}$, the recommendation results are denoted as $r_1,\cdots,r_{K_t}$, and the ground-truth words are denoted as $w_t^{(1)},\cdots,w_t^{(U)}$, which are words that aggregated from different users typing behavior. Please note that the recommender system we are considering is not a personalized one, therefore, it is reasonable to view each next word a user input in the context of $w_1,\cdots,w_{t-1}$ as a ground-truth word. The precision score is defined as:
\begin{equation*}
P@K=\frac{\sum_{u=1}^{U}\sum_{k=1}^{K}\delta(r_k=w_t^{(u)})}{\sum_{t=1}^{U}\delta(K^t>0)},
\end{equation*}
where $\delta(\cdot)$ is the indicator function and $K^t$ is the number of the recommendation words. That is, if $A$ is true, $\delta(A)=1$, otherwise $\delta(A)=0$.

Unlike precision that is the fraction of retrieved instances that are relevant, recall reflects the fraction of relevant instances that are retrieved:
\begin{equation*}
R@K=\frac{1}{U}\sum_{u=1}^{U}\sum_{k=1}^{K}\delta(r_k=w_t^{(u)}).
\end{equation*}
Since the cost of recommendation is much lower than the cost of input, this task value recall more important than precision. That is, the user prefer to glancing at the recommendation list rather than inputting the whole word on the mobile phone.

F1 both considers the precision and recall to compute the testing score, defined as follows.
\begin{equation*}
F1@K=\frac{P@K\times R@K}{\beta P@K+ (1-\beta)R@K}.
\end{equation*}

Regarding the Chinese input recommendation task as a ranking problem rather than a classification problem, it is suitable to use ranking measures for evaluation. In this paper, we take the mean average precision (MAP)\cite{CHEN:MAP,V:MAP2} as ranking measure for evaluation. The definitions are given as follows.
\begin{equation*}
MAP=\frac{1}{U}\sum_{u=1}^U\frac{1}{rank(w_t^{(u)})},
\end{equation*}
where $rank(w_t^{(u)})$ stands for the ranking position of $w_t^{(u)}$ in the recommendation list.

Specially for the task of Chinese input recommendation, we would like to evaluate from the perspective of how much the recommender system reduce the users' typing costs. Inspired by the measure of saved term defined in \cite{VS:SCSW}, we introduce two new evaluation measures in this paper, namely saved words and saved characters. The measure of the saved words($SW$) is the defined as percentage of words that users can directly select from the recommendation lists rather than typing directly on the mobile. While the measure of saved characters($SC$) further takes into account of the length of words, beyond the measure of saved words. We give the precise mathematical formulas as follows:
\begin{equation*}
SW=\frac{\sum_{u=1}^{U}\sum_{k=1}^{K}\delta(r_k=w_t^{(u)})}{U},
\end{equation*}
\begin{equation*}
SC=\frac{\sum_{u=1}^{U}\sum_{k=1}^{K}\delta(r_k=w_t^{(u)})lenc(w_t^{(u)})}{\sum_{u=1}^{U}lenc(w_t^{(u)})},
\end{equation*}
where $lenc(\cdot)$ stands for the length of the word ($\cdot$), i.e.~the number of characters contained in the word.

As we can see, the measure of $SW$ is the same as $R@K$. Therefore, the measure of Recall is more important than Precision in the Chinese input recommendation task, since the main concern of Chinese input recommender system is to rank the ground-truth words on top of the ranking list as most as possible so as to reduce users' input effort in the mobile devices. Considering the above issue, we set the $\beta$ in $F1$ measure as 2/3 in our experiments to emphasize the concern of Recall.

\section{Neural Language Models}
Though statistical language models such as n-gram based models have been widely used in the real Chinese input recommender systems and gain great success, it usually encounters serious sparsity problem because Chinese input recommendation task has some intrinsic characteristics.

\begin{CJK*}{UTF8}{gbsn}
(1) The representation of Chinese input can be quite diverse, even though users refer to the same meaning. For example, a chair in Chinese can be input as many forms, such as `凳子(stool)', `板凳(stool)', `椅子(chair)'. Since the n-gram have no generalization to other sequence of n words and no cross-generalization between different n-tuples, the n-gram has poor generalization and heavy sparse problem\cite{Bengio:DLNL}.

(2) Users' typing behaviors are quite different. Some users input Chinese sentences character by character, while others prefer the way of word by word, or even directly typing the Pinyin of the whole sentence.
\end{CJK*}

Though smoothed n-gram language models\cite{CS:SMOOTHMETHODS,ZCX:SMOOTH} can alleviate the sparsity problem\cite{TK:SPARSE}, there are at least two characteristics in this approach which beg to be improved upon. Firstly, it is not taking into account contexts father than 1 or 2 words \footnote{N-grams with up to 5 (i.e. 4 words of context) have been reported, though, but due to data scarcity, most predictions are made with a much shorter context}. Secondly, it does not take into account the `similarity' between words. Therefore, the effect of sparsity can not be well tackled with the smoothed statistical language models.

Recently, the approach of neural language models has been proposed to tackle these problems\cite{Z:DEEP,ZW:DEEP,CW:DEEP}. The idea can be summarized as follows. Firstly, each word in the vocabulary is associated with a distributed word feature vector (a real valued vector in $R^d$). Secondly, the joint probability of a word sequence is expressed based on the feature vectors of these words in the sequence. Thirdly, the word feature vectors and the parameters of the probability function is learned simultaneously. In this paper, we compare four popular neural language models, namely neural probabilistic language model (NLM) \cite{Bengio:FFNNLM}, word2vec \cite{MK:EMBEDDING,Mikolov:word2vec2}, recurrent neural network(RNN)\cite{MK:RNN} and long short term memory(LSTM)\cite{H:LSTM,G:LSTM2}.

\subsection{Probabilistic Neural Language Model}
Neural probabilistic language model (NLM for short) is proposed by Bengio et al. \cite{Bengio:FFNNLM}, which has been successfully used in many tasks\cite{embedding:app1,embedding:app2,embedding:app3}. Unlike n-grams only consider the estimation of conditional probability, the objective $f(w_1,\cdots,w_{t-n+1})=\hat{P}(w_t|w_1,\cdots,w_{t-1})$ in NLM is a composition of the two mappings $C$ and $g$. Specifically, $C$ is a mapping from each element $i\in V$ to a real vector $C(i)$, where $i$ is a word and $V$ is the vocabulary. The vector represents the distributed feature vector associated with each word in the vocabulary.
$g$ is a neural network with one hidden layer to map an input sequence of feature vectors for words in context, i.e.~$(C(w_{t-n+1}),\cdots,C(w_{t-1}))$, to the conditional probability for the next word $w_t$. The input vectors are first concatenate to form a word features layer activation vector $x=(C(w_{t-1}),C(w_{t-2}),\cdots,C(w_{t-n+1}))$. The hidden layer after an activation function will output $tanh(b_h+Hx)$, where $H$ is he hidden layer weights with $h$ hidden units, $b_h$ is the hidden bias, and  $tanh$ stands for the hyperbolic tangent activation function. The hidden layer output is further combined with the linear compositions of word features to form the input of softmax layer, i.e. $y=b+Wx+Utanh(b_h+Hx)$, where $U$ is the hidden-to-output weight, $W$ is the  weight of word features to output, and $b$ is the output bias. Finally, a softmax output layer is defined to output the probability as follows:
\begin{displaymath}
P(w_t|w_{t-1},\cdots,w_{t-n+1})=\frac{e^{y_{w_t}}}{\sum_i e^{y_i}},
\end{displaymath}
where $y_i$ stands for the score of word $i$.

We can see that the free parameters of this model can be expressed by $\theta=(b,b_h,W,U,H,C)$. They are all learned simultaneously. Training is achieved by looking for $\theta$ that maximizes the training corpus penalized log-likelihood:
\begin{displaymath}
L_n=\frac{1}{T}\sum_{i=1}^T \log f(w_t,\cdots,w_{t-n+1};\theta)+R(\theta),
\end{displaymath}
where $R(\theta)$ is a regularization term. For example, in NLM \cite{Bengio:FFNNLM}, $R$ is a weight decay penalty applied only to the weights of the neural network and to the $C$ matrix. Stochastic gradient ascent\cite{PL:PR} is usually used for optimization.

\subsection{Word2vec}
NLM is computationally expensive for training. Therefore, many more simple and efficient architecture is proposed, examples include \cite{MK:EMBEDDING,Mikolov:word2vec2,MK:RNN,BG:RNN,H:LSTM}. Among these methods, word2vec is the most successful one in terms of both efficiency and effective. Therefore, we choose it as the second representative neural language model.

Mikolov \cite{MK:EMBEDDING,Mikolov:word2vec2} proposed two new model architecture for learning distributed representations of words that try to minimize computational complexity, namely CBOW and Skip-Gram. These models both follow the paradigm that the continuous word vectors are learned using simple model, and the n-gram model is trained on top of these distributed vectors. When applying word2vec for language modeling, we can modify the above context $C(w_t)$ to indicate the previous $L$ words, i.e.~$C'(w_t)=(w_{t-L},\cdots,w_{t-1})$. The others are keep the same as the original word2vec. We can see that the corresponding architecture of CBOW and skip-gram can be modified to only consider half words in the context window. For simplicity, we still use the name `word2vec'. In this paper we only use CBOW for the our study, since it is more accordant with the language modeling task than skip-gram.

Specifically, given a word $w_t$ and its surrounding contexts $C'(w_t)=(w_{t-L},\cdots,w_{t-1})$, where $L$ stands for the size of the window, the conditional probability is given by:
\begin{equation*}
P(w_t|C'(w_t))=\frac{\exp\{v_{w_t}u(C'(w_t))\}}{\sum_{v\in V}\exp\{v_{w}u(C'(w_t))\}},
\end{equation*}
where $v_{w}$ stands for the word vector of each word $w\in V$, and $u(C'(w_t))$ is the context vector of word $w_t$. Specifically, $u(C'(w_t))$ can be sum, average, concatenate or max pooling of context word vectors, defined as follows.
\begin{displaymath}
u(C'(w_t))=F(v^c(w_{t-L}),\cdots,v^c(w_{t+L})),
\end{displaymath}
where $v^c(w)$ stands for the context vector of each word $w\in V$. In this paper, we use average as the function $F$, as that of word2vec tool.

Instead of using softmax for optimization, word2vec propose to use hierarchical softmax or negative sampling\cite{GB:word2vec,W2V:PMI} to save computation complexity. In this paper, we use the negative sampling approach. The basic idea is to
random select some words as the negative instances instead of all other words in language. Therefore, the log-likelihood function in CBOW can be represented as:
\begin{eqnarray*}
&L_c=\sum_{n=1}^N\sum_{w_t^n}[\log \sigma(v_{w_t^n}u(C'(w_t^n)))\\
&\quad+l E_{w'}\!\log\sigma(v_{w'}u(C'(w_t^n)))],
\end{eqnarray*}
where $w_t^n$ stands for the $t$-th word of the $n$-th sentence, $\sigma(x)=\frac{1}{1+e^{-x}}$, $l$ is the number of `negative' samples, $w'$ denotes the sampled negative words, which are sampled according to distribution $P_{nw}$. Stochastic gradient descent can be used for optimization, and the gradient is calculated via back-propagation algorithm.

\subsection{Recurrent Neural Network}
 \begin{CJK*}{UTF8}{gbsn}
Both NLM and word2vec have two shortcomings. (1)They directly compress the history to a single vector, without distinguishing the orders of different words in the history. However, the word order is usually crucial for this task. For example, the representations of `上天(up to the skies)’ and `天上(heaven)‘ are quite different. (2) They can not capture the long term dependencies of contexts, which is usually important for the next word prediction. For example, given a context `在(on) 海边(seaside) 旅游(travel), 我(I) 去(go to) 抓(catch)', the next word is `螃蟹(crab)' in the real data. We can see that the context `海边(seaside)' is crucial information for predicting the correct next word `螃蟹(crab)'. However, the NLM and word2vec are usually modeling the dependencies of contexts with a fixed-length window size, which may lose the long term dependencies.
 \end{CJK*}

To tackle this problem, Mikolov\cite{MK:RNN} used recurrent neural networks(RNN) as a language model. It can not only learn to compress the whole history to a low dimensional space, but also capture the sequence information and long term dependencies of the whole sentence. RNN first combines the current word embedding with the last-time context embedding(a hidden layer which accumulate all the context information) as an input. And secondly, updates the current context embedding. Finally, according to the current context, RNN will output the probability of the next word with a softmax function.

Specifically, the recurrent neural network consists of an input layer $x_{t-1}$, hidden layer $h_{t-1}$ and output layer $y_{t-1}$, where $x_{t-1}$ is the input to the network at time t-1, $y_{t-1}$ is the output and $h_{t-1}$ is the state of the network. At first, the input vector $x_{t-1}$ is defined as a combination of vector $C(w_{t-1})$ representing the current word and the hidden layer $h_{t-2}$ representing the historical context. That is to say, the network joints the history state at the late time t-2 and the current word representation as input to change the hidden state at current time t-1. Secondly, the hidden layer $h_{t-1}$ is represented as a function of the input vector $x_{t-1}$, where the hidden layer $h_{t-1}$ is set to be zero when t=1 is at the beginning of a sentence. Finally, the output layer $y_{t-1}$ is produced by the hidden state $h_{t-1}$. Mathematically, the recurrent neural network language model can be described as follows:
\begin{eqnarray*}
&x_{t-1}=&C(w_{t-1})+h_{t-2}\\
&h_{t-1}=&f(H_Rx_{t-1})\\
&y_{t-1}=&O_Rh_{t-1}
\end{eqnarray*}
where $H_R$ is the hidden matrix and $f$ is the sigmoid function to capture the non-linear information. $O_R$ is the output matrix. The output probability is a softmax function as:
\begin{displaymath}
P(w_t|w_{t-1},\cdots,w_1)=\frac{e^{y_{w_t}}}{\sum_i e^{y_i}}
\end{displaymath}

Given the previous context $w_1,\cdots,w_{t-1}$, the output layer $y_{t-1}$ represents the probability distribution of the next word $w_t$. At the training step, weights are updated by the backpropagation algorithm and the error is computed according to the cross entropy. That is to say, the log-likelihood function is defined as:
\begin{displaymath}
L_{t-1}=\vec{w}_t-\vec{y}_{t-1}
\end{displaymath}

\subsection{Long Short Term Memory}
Though RNN is capable to capture the long term dependencies of contexts, it usually faces the problem of gradient vanishing and gradient explosion for  long sentences \cite{H:LSTM,DIFFICULT:RNN}. The long short term memory(LSTM) is an advanced type of Recurrent Neural Network by using memory cells and gates to learn long term dependencies\cite{H:LSTM,G:LSTM2}. In LSTM, the cell stores the information state and the gates control whether the cell information should be transmited to other neural nodes or not.

The detailed formulation of LSTM is described as follows.
Given an input context ($w_0$,$w_1$,$\cdot$ ,$w_{t-1}$ ), where $C(w_t)$ is the word embedding of input word $w_t$ at time t. LSTM outputs a representation $h_t$ for position t as follows.
\begin{eqnarray*}
&i_t=&\sigma(W_{xi}C(w_t)+W_{hi}h_{t-1}+b_i)\\
&f_t=&\sigma(W_{xf}C(w_t)+W_{hf}h_{t-1}+b_f)\\
&c_t=&f_tc_{t-1}+i_ttanh(W_{xc}C(w_t)+W_{hc}h_{t-1}+b_c)\\
&o_t=&\sigma(W_{xo}C(w_t)+W_{ho}h_{t-1}+b_o)\\
&h_t=&o_ttanh(c_t)
\end{eqnarray*}
where $i$, $f$, $o$ denote the input, forget and output gates respectively, deciding whether the information of $C(w_t)$, $c_{t-1}$ and $c_t$ should be transmited or not . $c$ is the information stored in the memory cells and $h$ represents the probability distribution of the next word $w_t$.
The weights are updated by the backpropagation algorithm and the error is computed with the cross entropy. The log-likelihood function is defined as:
\begin{displaymath}
L_{t-1}=\vec{w}_t-\vec{h}_{t-1}
\end{displaymath}

\section{Empirical Settings}
We first describe the empirical settings in this section, including the Chinese input data and parameter settings of different models.

\begin{table}
\centering
\tbl{The distribution of word length on CIR.
\label{tb:distribution}
}
{
\begin{tabular}{ccccccc} \hline
word length & 1 & 2 & 3 & 4 & 5 & $>$5\\ \hline
frequency(\%) & 49.3 &37.2 &10.1 & 2.5 &0.6 & 0.3\\ \hline
\end{tabular}
}
\end{table}

\begin{table}
\centering
\caption{The distribution of sentence length on CIR.
\label{tb:Sdistribution}
}
{
\begin{tabular}{ccccccc} \hline
word length & 2 & 3 & 4 & 5 & $>$5\\ \hline
frequency(\%) & 27.1 & 20.3 & 14.5 & 10.3 & 27.4 \\ \hline
\end{tabular}
}
\end{table}
\subsection{Data Set}
\begin{CJK*}{UTF8}{gbsn}
In this paper, we collected a large scale data set from a commercial Chinese input recommendation(CIR) engine to facilitate our experiments.

The data set, named CIR, contains 20,000,000 word sequences. As we do not want to be biased by any artificial segmentation in the real application, we did not use any word segmentation tools to pre-process the data. Instead, we directly use the data segmented by the users themselves. For example, if a user type the word sequence `我们' as `wo(Pinyin of 我)'‘space’`men(Pinyin of 们)', `我' and `们' can be viewed as two distinct words. Otherwise if another use type the word sequence `我们' as `women(Pinyin of 我们)'， the word `我们' can be viewed as a single word.
\end{CJK*}

We made a statistic of the number of Chinese characters in a word on the CIR data, shown in Table \ref{tb:distribution}. We can see that each word contains 1.7 characters on average, and almost 49.3\% of the words in the corpus are with the length of one character, which means the majority of users prefer to inputting single word. So the n-gram, counting the co-occurrence frequency, adds bias to predict the most frequent single words based on the same preceding context.

We also made a statistics of the number of words in a sequence on the CIR data, shown in Table \ref{tb:Sdistribution}. The results show that each sequence has 4.9 words on average, and the sequences are with various length.

 \begin{CJK*}{UTF8}{gbsn}
To facilitate the empirical study of language models for Chinese input recommendation, the CIR corpus has been preprocessed by replacing all numbers with one notation `NUM' and ignoring the English words.
Furthermore, the word sequence can be constructed to many (context, predicted word) pairs as follows: Firstly, each subset of previous words can be viewed as a context, and the following word is viewed as the predicted word. Therefore, a word sequence with length $n$ can be constructed to $n-1$ (context, predicted word) pairs. Secondly, pairs with the same context can be aggregated together for the language modeling task.For example, given two (context, predicted word) pairs (`谢谢(Thanks)你的(your)’, ` 招待(treat)‘) and (`谢谢(Thanks)你的(your)’, `邀请(invitation)‘), we can obtain the following data by aggregating: i.e.~the context is`谢谢你的(Thanks your)', the groundtruth of predicted words are `招待(treat)’ and `邀请(invitation)'. We can see that such data are suitable for the language modeling task after the preprocessing. For our study, we randomly select 80{\%} sequences from the whole CIR data as training set, 10{\%} sequences as validation set and the rest 10{\%} as test set.
\end{CJK*}

\pgfplotsset{
axis background/.style={fill=mercury},
grid=both,
  xtick pos=left,
  ytick pos=left,
  tick style={
    major grid style={style=white,line width=1pt},
    minor grid style=gallery,
    draw=none
    },
  minor tick num=1,
}
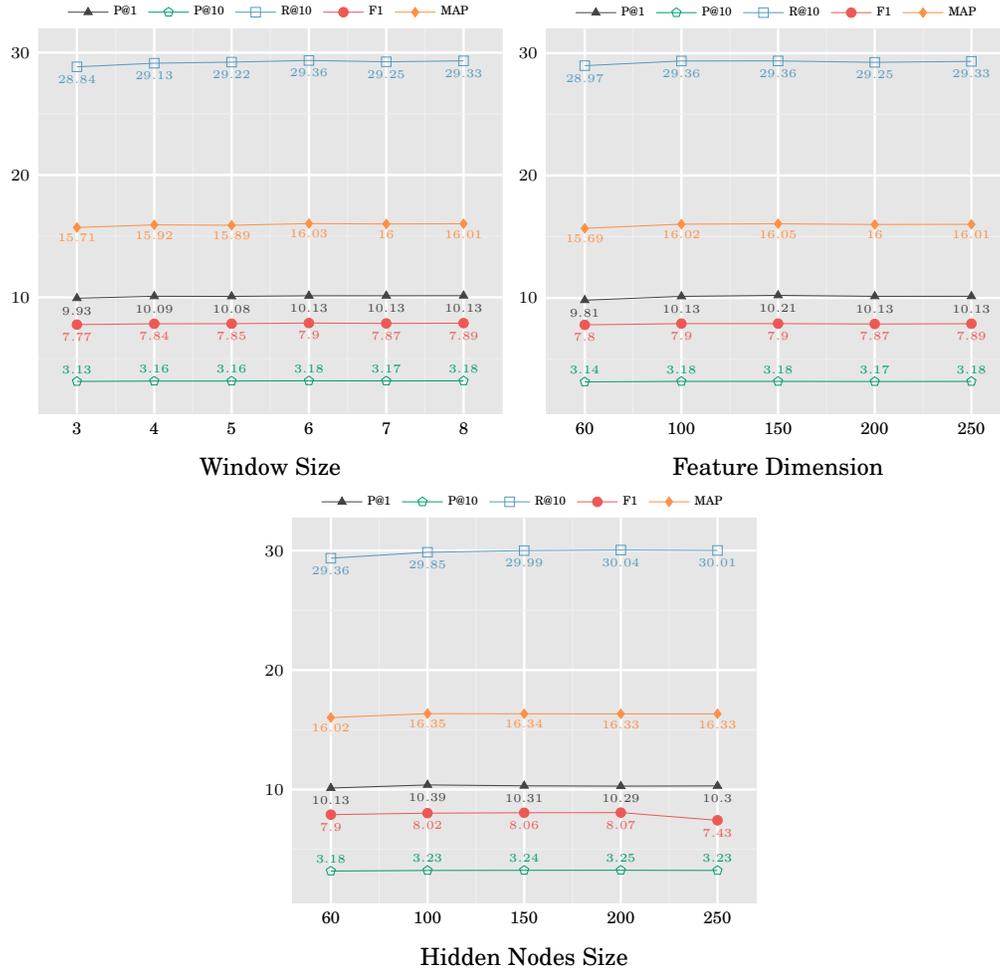
\begin{figure}
\centering
  \begin{tikzpicture}[scale=0.9]
    \begin{axis}[
      xlabel=Window Size,
      xticklabels={3,4,5,6,7,8},
      xtick={1,2,3,4,5,6},
      legend style={
          font=\tiny,
          draw=none,
          legend columns=-1,
          at={(0.5,1)},
          anchor=south,
          /tikz/every even column/.append style={column sep=0.9mm}
        },
        ymajorgrids,
        major grid style={draw=white},
        y axis line style={opacity=0},
        tickwidth=0pt,
        nodes near coords,
        every node near coord/.append style={anchor=north, font=\fontsize{4pt}{4pt}\selectfont},
        legend entries = {P@1, P@10,R@10,F1,MAP},
        every tick label/.append style={font=\scriptsize}
        ]
       \addplot[color=tuatara,mark=triangle*] coordinates {
(1,9.9289)
(2,10.0884)
(3,10.0788)
(4,10.1262)
(5,10.1299)
(6,10.1339)
      };
      \addplot[color=free_speech_aquamarine, mark=pentagon, every node near coord/.append style={anchor=south}] coordinates {
(1,3.1302)
(2,3.1573)
(3,3.1590)
(4,3.1812)
(5,3.1684)
(6,3.1769)
      };

      \addplot[color=shakespeare,mark=square] coordinates {
(1,28.8351)
(2,29.1310)
(3,29.2204)
(4,29.3568)
(5,29.2473)
(6,29.3251)
      };
      \addplot[color=flamingo, mark=*, every node near coord/.append style={anchor=north}] coordinates {
(1,7.7724)
(2,7.8419)
(3,7.8498)
(4,7.9016)
(5,7.8701)
(6,7.8912)
      };
	\addplot[color=sun_shade, mark=diamond*, every node near coord/.append style={yshift=0.3mm,anchor=north}] coordinates {
(1,15.7072)
(2,15.9230)
(3,15.8903)
(4,16.0254)
(5,15.9970)
(6,16.0114)
      };
    \end{axis}
  \end{tikzpicture}
 \begin{tikzpicture}[scale=0.9]
    \begin{axis}[
      xlabel=Feature Dimension,
      xticklabels={60,100,150,200,250},
      xtick={1,2,3,4,5},
      legend style={
          font=\tiny,
          draw=none,
          legend columns=-1,
          at={(0.5,1)},
          anchor=south,
          /tikz/every even column/.append style={column sep=0.9mm}
        },
        ymajorgrids,
        major grid style={draw=white},
        y axis line style={opacity=0},
        tickwidth=0pt,
        nodes near coords,
        every node near coord/.append style={anchor=north, font=\fontsize{4pt}{4pt}\selectfont},
        legend entries = {P@1, P@10,R@10,F1,MAP},
        every tick label/.append style={font=\scriptsize}
        ]
       \addplot[color=tuatara,mark=triangle*] coordinates {
(1,9.8127)
(2,10.1251)
(3,10.2064)
(4,10.1299)
(5,10.1339)
      };
      \addplot[color=free_speech_aquamarine, mark=pentagon, every node near coord/.append style={anchor=south}] coordinates {
(1,3.1400)
(2,3.1809)
(3,3.1806)
(4,3.1684)
(5,3.1769)
      };

      \addplot[color=shakespeare,mark=square] coordinates {
(1,28.9732)
(2,29.3563)
(3,29.3620)
(4,29.2473)
(5,29.3251)
      };
      \addplot[color=flamingo, mark=*, every node near coord/.append style={anchor=north}] coordinates {
(1,7.7990)
(2,7.9009)
(3,7.9006)
(4,7.8701)
(5,7.8912)
      };
	\addplot[color=sun_shade, mark=diamond*, every node near coord/.append style={yshift=0.3mm,anchor=north}] coordinates {
(1,15.6851)
(2,16.0242)
(3,16.0545)
(4,15.9970)
(5,16.0114)
    };
    \end{axis}
  \end{tikzpicture}
  \begin{tikzpicture}[scale=0.9]
    \begin{axis}[
      xlabel=Hidden Nodes Size,
      xticklabels={60,100,150,200,250},
      xtick={1,2,3,4,5},
      legend style={
          font=\tiny,
          draw=none,
          legend columns=-1,
          at={(0.5,1)},
          anchor=south,
          /tikz/every even column/.append style={column sep=0.9mm}
        },
        ymajorgrids,
        major grid style={draw=white},
        y axis line style={opacity=0},
        tickwidth=0pt,
        nodes near coords,
        every node near coord/.append style={anchor=north, font=\fontsize{4pt}{4pt}\selectfont},
        legend entries = {P@1, P@10,R@10,F1,MAP},
        every tick label/.append style={font=\scriptsize}
        ]
       \addplot[color=tuatara,mark=triangle*] coordinates {
(1,10.1251)
(2,10.3854)
(3,10.3110)
(4,10.2857)
(5,10.3027)
      };
      \addplot[color=free_speech_aquamarine, mark=pentagon, every node near coord/.append style={anchor=south}] coordinates {
(1,3.1809)
(2,3.2299)
(3,3.2427)
(4,3.2479)
(5,3.2311)
      };

      \addplot[color=shakespeare,mark=square] coordinates {
(1,29.3563)
(2,29.8525)
(3,29.9944)
(4,30.0366)
(5,30.0068)
      };
      \addplot[color=flamingo, mark=*, every node near coord/.append style={anchor=north}] coordinates {
(1,7.9009)
(2,8.0247)
(3,8.0576)
(4,8.0703)
(5,7.4276)
      };
	\addplot[color=sun_shade, mark=diamond*, every node near coord/.append style={yshift=0.3mm,anchor=north}] coordinates {
(1,16.0242)
(2,16.3508)
(3,16.3374)
(4,16.3292)
(5,16.3297)
      };
    \end{axis}
  \end{tikzpicture}
  \caption{\label{fig:windowdimensionhiddenNLM} Influences of window size, feature dimension and hidden nodes size for NLM.}
\end{figure}
\subsection{Parameter Settings}
Now we introduce some parameter settings in n-grams, NLM, RNN, LSTM and word2vec.

We adopt two statistical language models for comparison, pure n-grams (n-gram for short), and n-grams with interpolated Kneser-Ney smoothing (n-gram-KN for short)\cite{Bengio:FFNNLM}. For the two models, we use unigram, bigram and trigram for counting. 

For NLM, we make an experiment on the validation data of CIR to test the different window size, embedding dimensions and different number of hidden nodes. The performance results of P@1,10,R@10,F1@10,MAP with different window size, different embedding dimension and hidden nodes size are reported in Figure \ref{fig:windowdimensionhiddenNLM}. Specifically, 
we first set the feature dimension as 100 and the hidden nodes as 60, and test the window size from $3$ to $8$ with step $1$. We can see that the performances first increase and keep stable as the window size increase. Because of the average sentence length about 4.9 words, the window size more than 6 is good enough for these task. 
Secondly, we set the window size as 6 and the hidden nodes as 60 to test the embedding dimension as $60$,$100$,$150$,$200$,$250$. The performance results first increase and then keep stable as the feature dimension increase. Therefore, the feature dimension is set to 100, and we further test the hidden nodes of $60$,$100$,$150$,$200$,$250$ with the window size set to 6. We can see that the performances will be improved significantly as the hidden nodes increase, which means that more hidden nodes can represent the meaning more detailed.  
Considering the computational cost, the dimension of word embedding, window size and the number of hidden units is set to be $100$, $6$, and $200$ in the latter experiments.

\begin{figure}
\centering
 \begin{tikzpicture}[scale=0.9]
    \begin{axis}[
      xlabel=Feature Dimension,
      xticklabels={60,100,150,200,300},
      xtick={1,2,3,4,6},
      legend style={
          font=\tiny,
          draw=none,
          legend columns=-1,
          at={(0.5,1)},
          anchor=south,
          /tikz/every even column/.append style={column sep=0.9mm}
        },
        ymajorgrids,
        major grid style={draw=white},
        y axis line style={opacity=0},
        tickwidth=0pt,
        nodes near coords,
        every node near coord/.append style={anchor=north, font=\fontsize{4pt}{4pt}\selectfont},
        legend entries = {P@1, P@10,R@10,F1,MAP},
        every tick label/.append style={font=\scriptsize}
        ]
       \addplot[color=tuatara,mark=triangle*, every node near coord/.append style={anchor=south}] coordinates {
(1,8.4572)
(2,8.4942)
(3,8.5107)
(4,8.5471)
(6,8.2721)
      };
      \addplot[color=free_speech_aquamarine, mark=pentagon, every node near coord/.append style={anchor=south}] coordinates {
(1,2.9551)
(2,2.9874)
(3,2.9980)
(4,3.0025)
(6,2.9779)
      };

      \addplot[color=shakespeare,mark=square,every node near coord/.append style={anchor=north}] coordinates {
(1,27.1726)
(2,27.5003)
(3,27.6052)
(4,27.7101)
(6,27.4872)
      };
      \addplot[color=flamingo, mark=*, every node near coord/.append style={anchor=north}] coordinates {
(1,7.3352)
(2,7.4170)
(3,7.4435)
(4,7.4579)
(6,7.3970)
      };
	\addplot[color=sun_shade, mark=diamond*, every node near coord/.append style={yshift=0.3mm,anchor=north}] coordinates {
(1,14.1199)
(2,14.2579)
(3,14.2958)
(4,14.3337)
(6,13.9987)
      };
    \end{axis}
  \end{tikzpicture}
  \caption{\label{fig:dimensionhiddenRNN} Influences of feature dimension and hidden nodes size for RNN.}
\end{figure}
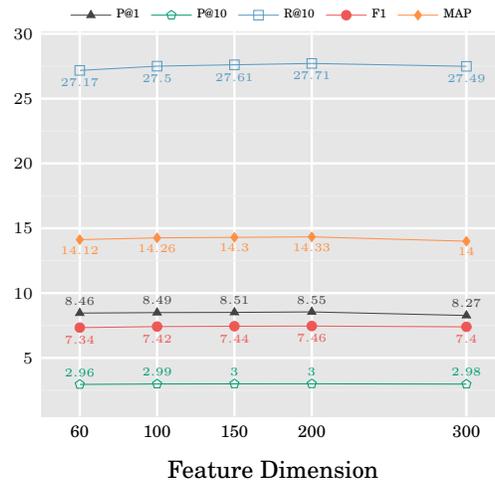

For RNN, we test the different embedding dimensions and different number of hidden nodes on the validation set. Specifically, we set embedding dimension and hidden nodes the same, and test them as $60$,$100$,$150$,$200$,$300$. The performances of embedding dimension and different number of hidden nodes are reported in Figure \ref{fig:dimensionhiddenRNN}. As we can see, the P@1,10,R@10,F1 and MAP first increase and then drop as the growth of feature dimension and hidden nodes size. Therefore, the dimension of word vector and the number of hidden unite are both set to $200$.

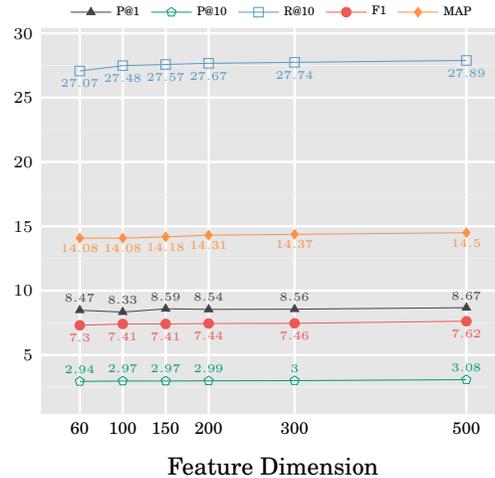
\begin{figure}
\centering
  \begin{tikzpicture}[scale=0.9]
    \begin{axis}[
      xlabel=Feature Dimension,
      xticklabels={60,100,150,200,300,500},
      xtick={1,2,3,4,6,10},
      legend style={
          font=\tiny,
          draw=none,
          legend columns=-1,
          at={(0.5,1)},
          anchor=south,
          /tikz/every even column/.append style={column sep=0.9mm}
        },
        ymajorgrids,
        major grid style={draw=white},
        y axis line style={opacity=0},
        tickwidth=0pt,
        nodes near coords,
        every node near coord/.append style={anchor=north, font=\fontsize{4pt}{4pt}\selectfont},
        legend entries = {P@1, P@10,R@10,F1,MAP},
        every tick label/.append style={font=\scriptsize}
        ]
       \addplot[color=tuatara,mark=triangle*, every node near coord/.append style={anchor=south}] coordinates {
(1,8.4739)
(2,8.3302)
(3,8.5862)
(4,8.5407)
(6,8.5571)
(10,8.6721)
      };
      \addplot[color=free_speech_aquamarine, mark=pentagon, every node near coord/.append style={anchor=south}] coordinates {
(1,2.9419)
(2,2.9736)
(3,2.9725)
(4,2.9880)
(6,3.0025)
(10,3.0779)
      };

      \addplot[color=shakespeare,mark=square] coordinates {
(1,27.0670)
(2,27.4839)
(3,27.5739)
(4,27.6652)
(6,27.7401)
(10,27.8872)
      };
      \addplot[color=flamingo, mark=*, every node near coord/.append style={anchor=north}] coordinates {
(1,7.3032)
(2,7.4083)
(3,7.4115)
(4,7.4435)
(6,7.4579)
(10,7.6196)
      };
	\addplot[color=sun_shade, mark=diamond*, every node near coord/.append style={yshift=0.3mm,anchor=north}] coordinates {
(1,14.0799)
(2,14.0786)
(3,14.1799)
(4,14.3058)
(6,14.3737)
(10,14.498)
      };
    \end{axis}
  \end{tikzpicture}
  \caption{\label{fig:dimensionhiddenLSTM} Influences of feature dimension and hidden nodes size for LSTM.}
\end{figure}

For LSTM, we test the different embedding dimensions and different number of hidden nodes on the validation set. We choose embedding dimension and hidden nodes from $60$,$100$,$150$,$200$,$300$, and $500$. We report the performances of different feature dimensions and different number of hidden nodes in Figure \ref{fig:dimensionhiddenLSTM}. The trend of evaluation measure is the same as that of RNN. Therefore, the dimension of word vector and the number of hidden unite are also set to $300$.

\begin{figure}
\centering
  \begin{tikzpicture}[scale=0.9]
    \begin{axis}[
      xlabel=Window Size,
      xticklabels={1,2,3,4,5,6,7,8},
      xtick={1,2,3,4,5,6,7,8},
      legend style={
          font=\tiny,
          draw=none,
          legend columns=-1,
          at={(0.5,1)},
          anchor=south,
          /tikz/every even column/.append style={column sep=0.9mm}
        },
        ymajorgrids,
        major grid style={draw=white},
        y axis line style={opacity=0},
        tickwidth=0pt,
        nodes near coords,
        every node near coord/.append style={anchor=north, font=\fontsize{4pt}{4pt}\selectfont},
        legend entries = {P@1, P@10,R@10,F1,MAP},
        every tick label/.append style={font=\scriptsize}
        ]
       \addplot[color=tuatara,mark=triangle*, every node near coord/.append style={anchor=south}] coordinates {
(1,5.692677)
(2,6.350747)
(3,6.306719)
(4,6.162821)
(5,6.249130)
(6,6.353259)
(7,6.423931)
(8,6.460067)
      };
      \addplot[color=free_speech_aquamarine, mark=pentagon, every node near coord/.append style={anchor=south}] coordinates {
(1,1.713246)
(2,2.068650)
(3,2.128329)
(4,2.124484)
(5,2.150950)
(6,2.166821)
(7,2.178636)
(8,2.187481)
      };

      \addplot[color=shakespeare,mark=square] coordinates {
(1,15.247386)
(2,18.407471)
(3,18.932329)
(4,18.895629)
(5,19.131771)
(6,19.285457)
(7,19.400757)
(8,19.483186)
      };
      \addplot[color=flamingo, mark=*, every node near coord/.append style={anchor=north}] coordinates {
(1,4.2273)
(2,5.1041)
(3,5.2510)
(4,5.2414)
(5,5.3067)
(6,5.3465)
(7,5.3762)
(8,5.3982)
      };
	\addplot[color=sun_shade, mark=diamond*, every node near coord/.append style={yshift=0.3mm,anchor=north}] coordinates {
(1,8.561830)
(2,9.940703)
(3,10.031594)
(4,9.889473)
(5,10.016753)
(6,10.148334)
(7,10.235289)
(8,10.281117)
      };
    \end{axis}
  \end{tikzpicture}
  \begin{tikzpicture}[scale=0.9]
    \begin{axis}[
      xlabel=Feature dimension,
      xticklabels={50,100,150,200,250},
      xtick={1,2,3,4,5},
      legend style={
          font=\tiny,
          draw=none,
          legend columns=-1,
          at={(0.5,1)},
          anchor=south,
          /tikz/every even column/.append style={column sep=0.9mm}
        },
        ymajorgrids,
        major grid style={draw=white},
        y axis line style={opacity=0},
        tickwidth=0pt,
        nodes near coords,
        every node near coord/.append style={anchor=north, font=\fontsize{4pt}{4pt}\selectfont},
        legend entries = {P@1, P@10,R@10,F1,MAP},
        every tick label/.append style={font=\scriptsize}
        ]
       \addplot[color=tuatara,mark=triangle*, every node near coord/.append style={anchor=south}] coordinates {
(1,5.99133)
(2,6.21807)
(3,6.25388)
(4,6.31334)
(5,6.31140)
      };
       \addplot[color=free_speech_aquamarine, mark=pentagon, every node near coord/.append style={anchor=south}] coordinates {
(1,2.09870)
(2,2.13547)
(3,2.16172)
(4,2.20909)
(5,2.23533)
      };

      \addplot[color=shakespeare,mark=square] coordinates {
(1,18.62389)
(2,19.01733)
(3,19.26639)
(4,19.67494)
(5,19.91491)
      };
      \addplot[color=flamingo, mark=*, every node near coord/.append style={anchor=north}] coordinates {
(1,5.1756)
(2,5.2697)
(3,5.3353)
(4,5.4515)
(5,5.5166)
      };
	\addplot[color=sun_shade, mark=diamond*, every node near coord/.append style={yshift=0.3mm,anchor=north}] coordinates {
(1,9.73016)
(2,9.96935)
(3,10.04341)
(4,10.18101)
(5,10.22354)
      };
    \end{axis}
  \end{tikzpicture}
  \begin{tikzpicture}[scale=0.9]
    \begin{axis}[
      xlabel=Negative Samples,
      xticklabels={1,3,5,7,10,15,20},
      xtick={1,3,5,7,10,15,20},
      legend style={
          font=\tiny,
          draw=none,
          legend columns=-1,
          at={(0.5,1)},
          anchor=south,
          /tikz/every even column/.append style={column sep=0.9mm}
        },
        ymajorgrids,
        major grid style={draw=white},
        y axis line style={opacity=0},
        tickwidth=0pt,
        nodes near coords,
        every node near coord/.append style={anchor=north, font=\fontsize{4pt}{4pt}\selectfont},
        legend entries = {P@1, P@10,R@10,F1,MAP},
        every tick label/.append style={font=\scriptsize}
        ]
       \addplot[color=tuatara,mark=triangle*, every node near coord/.append style={anchor=south}] coordinates {
(1,4.17093)
(3,6.24913)
(5,5.51327)
(7,4.80395)
(10,4.10684)
(15,3.59980)
(20,3.38171)
      };
       \addplot[color=free_speech_aquamarine, mark=pentagon, every node near coord/.append style={anchor=south}] coordinates {
(1,1.85393)
(3,2.15095)
(5,1.90756)
(7,1.75038)
(10,1.58107)
(15,1.46090)
(20,1.41864)
      };

      \addplot[color=shakespeare,mark=square] coordinates {
(1,16.28469)
(3,19.13177)
(5,17.00457)
(7,15.61650)
(10,14.09026)
(15,12.96667)
(20,12.58034)
      };
      \addplot[color=flamingo, mark=*, every node near coord/.append style={anchor=north}] coordinates {
(1,4.5632)
(3,5.3067)
(5,4.7082)
(7,4.3209)
(10,3.9021)
(15,3.6028)
(20,3.4980)
      };
	\addplot[color=sun_shade, mark=diamond*, every node near coord/.append style={yshift=0.3mm,anchor=north}] coordinates {
(1,7.50891)
(3,10.01675)
(5,8.74614)
(7,7.70833)
(10,6.65142)
(15,5.88275)
(20,5.58384)
      };
    \end{axis}
  \end{tikzpicture}
  \caption{\label{fig:windowdimensionnegativeword2vec} Influences of feature dimension and negative samples for word2vec.}
\end{figure}
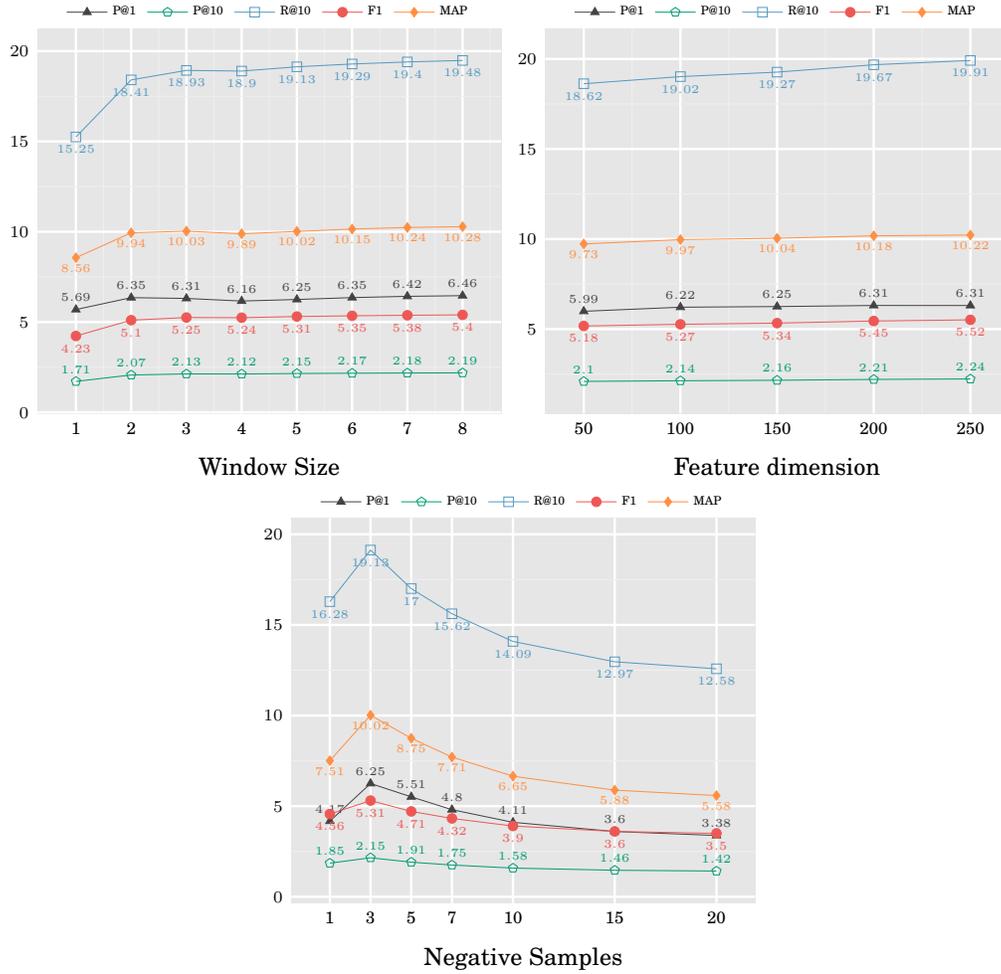

For word2vec, we test the influences of different window sizes, feature dimensions and the different number of negative samples on the validation set of CIR. Firstly, the feature dimension is set to 100, and we test different window sizes varying from 1 to 9 with step 1. The P@1,10, R@10,F1 and MAP scores with different settings are reported in Figure~\ref{fig:windowdimensionnegativeword2vec}. The results show that the P@10,R@10 and F1 will be improved as the window size increases, and the P@1 and MAP reach the top with the window size set to 2, and then keeps stable when the window size is larger than 5. Based on these observations, we set window size to 5 in the following experiments.
After that, we study the influence of different feature dimensions, i.e.~50,100,150,200,250. The experimental results are shown is Figure~\ref{fig:windowdimensionnegativeword2vec}.
We can see that the performance keeps stable among P@1,10, F1 and MAP. While for R@10, the performance first increase then drop, with the growth of the feature dimension. Therefore, we set the feature dimension as 200 in our experiments. This indicates that word2vec needs an appropriate feature dimension for representing a word, so that the true `similar' words can be retrieved. As for why the precision is not sensitive to the feature dimension, the advantage of word2vec lies in providing some more similar word to give some reasonable choices, but not in predicting the accurate words.
For the number of negative samples, we set 1,3,5,7,10,15,20 as the number of negative samples and report the results in Figure~\ref{fig:windowdimensionnegativeword2vec}. The experimental results show that the performance will consistently first increase then drop when the negative samples is larger than 3. So we set the number of negative samples as 3 in our experiments.

\begin{table*}
\centering
\tbl{
The comparison results among statistical, neural and hybrid language models.
\label{tb:combination}
}
{
\begin{tabular}{ccccccccc} \hline
model & P@1&P@3 &P@5 &P@10& R@10& F1 & MAP & SC \\ \hline
ngram & \underline{12.713} & 	\underline{7.289} & 	\underline{5.369} & 	3.414 & 	\underline{29.766} & 	\underline{8.391} & 	\underline{18.089} & 	\underline{29.596}   \\
ngram-KN & 9.804 & 	6.842 & 	4.971 & 	\underline{3.542} & 	28.775 & 	7.743 & 	15.314 & 	28.564   \\ \hline
NLM & 10.126 & 	6.472 & 	4.892 & 	3.181 & 	29.357 & 	7.902 & 	16.025 & 	29.226   \\
word2vec & 6.313 & 	4.073 & 	3.120 & 	2.209 & 	19.675 & 	5.452 & 	10.181 & 	19.454  \\
RNN & 8.547 & 	5.796 & 	4.509 & 	3.002 & 	27.710 & 	7.417 & 	14.334 & 	27.586 \\
LSTM & 8.672 & 	5.766 & 	4.517 & 	3.078 & 	27.887 & 	7.381 & 	14.499 & 	27.363  \\ \hline\hline
NLM+ngram& 13.154 & 	7.536 & 	5.554 & 	3.467 & 	30.683 & 	8.545 & 	18.777 & 	30.508  \\
word2vec+ngram& 12.938 & 	7.451 & 	5.499 & 	3.487 & 	30.483 & 	8.576 & 	18.473 & 	30.312  \\
weighted+ngram& 13.126 & 	7.618 & 	5.522 & 	3.499 & 	30.587 & 8.605 & 	18.545 & 30.471\\
RNN+ngram& 12.981 & 	7.429 & 	5.479 & 	3.483 & 	30.412 & 	8.563 & 	18.468 & 	30.234   \\
LSTM+ngram& 12.986 & 	7.430 & 	5.470 & 	3.474 & 	30.333 & 	8.542 & 	18.457 & 	30.156  \\
NLM+word2vec+ngram& {\bf 14.979} & 	{\bf 7.821} & 	{\bf  6.34} & 	{\bf  3.891 }& {\bf  31.18} & {\bf  9.407} & {\bf 19.587} & {\bf  30.938} \\ \hline
\end{tabular}
}
\end{table*}
\section{Experimental Results}\label{sec:experiments}
In this section, we show our experimental results on the Chinese input recommendation task. Firstly,  we conducted experiments to study the comparison of different approaches. Secondly, we combined the statistical and neural language models, and found that better results can be obtained than using a single one. Thirdly, we conducted a further discussion on different neural methods to provide some more insights.

\subsection{Comparison Between Statistical and Neural Language Models}
We compare all the models on CIR for fair comparison, and the experimental results are shown in Table~\ref{tb:combination}. From the results, we can see that two statistical methods (i.e.~n-gram and n-gram-KN) obtain comparable results. Similar results can be obtained for the four neural methods (i.e.~NLM,  word2vec, RNN and LSTM). As for the comparisons between statistical and neural approaches, we can see that the statistical language models performs much better than that of the neural ones. Taking MAP as an example, the best statistical language model (i.e.~n-gram/18.09) can improve the best neural language model (i.e.~NLM/16.03) by 12.85\%. An exception is R@10, we can see that NLM performs slightly better than n-gram. We make a further analysis to show the differences between statistical and neural language models.
\begin{table}
\centering
\tbl{
The sparsity rate of n-gram and NLM with different number and average length of context words.
\label{tb:sparse}
}
{
\begin{tabular}{ccccccc} \hline
number & ngram & neural &  & average length & ngram & neural\\ \hline
1 & 23.38 & 1.58 &   & [1,2) & 1.04 & 0.001\\
2 & 6.51 & 0.01 &   & [2,3) & 7.42 & 0.03\\
3 & 4.07 & 0.001  &   & [3,4) & 25.54 & 0.36\\
4 & 3.49 & 0.0002 &   & [4,5) & 41.27 & 1.31\\
5 & 2.96 & 0  &   & [5,) & 52.2 & 2.02\\ \hline
\end{tabular}
}
\end{table}

\subsubsection{Sparsity Rate Analysis}
~~\\
Firstly, we give some analysis to show how serious the sparsity problem is for statistical and neural language models, and explain why this happens. Specifically, we define a measure named sparsity rate as $sr=\frac{\#\{no-recommendation\}}{\#\{chances-to-recommend\}}$. We made a statistics on the sparsity rate for different models, and show the results on Table~\ref{tb:sparse}. Since the statistics for neural language models are similar, we only show the results of NLM for demonstration.

\begin{CJK*}{UTF8}{gbsn}
At first, we conduct an analysis on the influence of sparsity rate by the number of context words. We test different number of context words from 1 to 5 with step 1, and show the $sr$ in the left of the table. From the results, we can see that the sparsity rate of n-gram is about 23.4\% when the context contains only one word. And it drops significantly when the context contains more words. The reason is that it can only use low-order n-gram when the context is short. For example, the context is `凳子上(on stool)' but the training data only contain bi-grams of `椅子上(on chair)+睡(sleep)' and `椅子上(on char)+的('s)'. However, the $sr$ will drops for neural language models and the largest one is only $1.58\%$. That is because the neural language model can seize the similarities between words(ie.~凳子stool and 椅子chair), and thus largely improve the sparsity problem than statistical language model.

Secondly, we test the average length of context from $[1,2),[2,3),\cdots,[5,\infty)$, and show the $sr$ in the right of the table. As we can see, the $sr$ increases as the  average length increases for both statistical and neural language models. Specifically, statistical language model can reach to almost 52.2\% when the average length is more than 5 chinese characters, which means that the input behaviors can affect the sparse rate directly. For the same meaning `我明天要去超市(I tomorrow will go to supermarket)', some users may input `我明天要去(I tomorrow will go to)' as his first input and other users may input `我(I)'`明天(tomorrow)'`要去(will go to)' step by step. The average length of context for the former user is 5 and the last user is 1.67. From our experimental results, n-gram can not performs well for the former users. While for the neural language model, the sparsity problem can be alleviated by leveraging semantics between different words, and the largest $sr$ is about 2.02\% for NLM, which is much smaller than the n-gram based statistical language model.
\end{CJK*}

\begin{table}
\centering
\tbl{
The overlapping rates of the recommendation results between each two models.
\label{tb:overlap}
}
{
\begin{tabular}{ccccccc} \hline
overlap & ngram &ngram-KN & NLM & word2vec  & RNN & LSTM\\ \hline
ngram &1 & 0.590726 & 0.1855 & 0.143099 & 0.198572 & 0.200481\\
ngram-KN &  & 1 & 0.201453 & 0.145458 & 0.224591 & 0.228102\\
NLM & &  & 1 & 0.0858913 & 0.532115 & 0.525754\\
word2vec &  &  &  & 1 & 0.0794111 & 0.0792779\\
RNN & &  &  &  & 1 & 0.727284\\
LSTM & & & &  &  &1\\  \hline
\end{tabular}
}
\end{table}
\subsubsection{Overlapping Rate Analysis}
~~\\
Secondly, we conduct a qualitative analysis between n-gram and neural language models. Specifically, we make a statistics on the overlapping rates of the recommendation results between each two models, among n-gram, n-gram-KN, NLM, word2vec, RNN and LSTM. The experimental results are shown in Table~\ref{tb:overlap}. From the results, we can first see that there is large overlap between n-gram and n-gram-KN (i.e.~59.07\%). This is understandable since n-gram-KN is just a smoothed version of n-gram. Secondly, there is large overlap rate among some neural language models, such as NLM, RNN and LSTM. For example RNN and NLM has 53.21\% overlap, LSTM and NLM(i.e.~52.58\%), RNN and LSTM(i.e.~72.73\%). All these neural language models may predict the similar results. However, the word2vec is different from other neural language models, since the overlapping rate between word2vec and others is quite small, i.e.~$8.5\%, 7.9\%$ and $7.9\%$, respectively. This may be caused by the negative sampling strategy used in word2vec. For both NLM and RNN, the likelihood function is utilized as the loss function. That is to say, the words which is popular will be encouraged. However, they will be penalized in word2vec when using negative sampling. Thirdly, there is small overlap between statistical and neural language models. For example, the overlap between n-gram and NLM is 18.55\%, while that between n-gram and word2vec is 14.31\%. Therefore, statistical and neural language models will provide different recommendations. These results indicate that we can combine different approaches to obtain a better results. We will give further investigations on this issue later.

\subsubsection{Case Studies}
~~\\
\begin{CJK*}{UTF8}{gbsn}
\begin{table*}
\centering
\tbl{\label{tab:case} The case studies on the comparisons of statistical and neural language models.}
{
\newcommand{\tabincell}[2]{\begin{tabular}{@{}#1@{}}#2\end{tabular}}

\begin{tabular}{cccccc} \hline
ngram & ngram-KN & NLM & word2vec & RNN & LSTM \\ \hline
\tabincell{c}{电话(phone)\\了\\的\\你(your)\\我(my)} & \tabincell{c}{电话(phone)\\的\\了\\NUM\\你(your)} & \tabincell{c}{电话(phone)\\疫苗(vaccine)\\针(have an injection)\\交道(contact)\\点滴(transfuse)} & \tabincell{c}{羽毛球(badminton)\\预防针(vaccine)\\小报告(report)\\麻将(mahjong)\\游戏(games)} & \tabincell{c}{电话(phone)\\NUM\\级(upgrade)\\没人接(no answer)\\短信(message)} & \tabincell{c}{电话(phone)\\工(work) \\牌(cards) \\麻将(mahjong) \\你(you) }\\ \hline
\tabincell{c}{号(ID)\\上(sign in)\\红包(money)\\了\\群(group)}& \tabincell{c}{是(is)\\的\\好友(friend)\\群(group)\\红包(money)} & \tabincell{c}{好友(friend)\\群(group)\\支付(pay)\\红包(money)\\你(you)} & \tabincell{c}{密码(password)\\账号(ID)\\下载(download)\\支付(pay)\\转账(virement)} & \tabincell{c}{密码(password)\\NUM\\给你(give you)\\付款(pay)\\下单(order)}  & \tabincell{c}{ 号(ID)\\了 \\发(give) \\NUM \\红包(money)} \\ \hline
\end{tabular}
}
\end{table*}
Table~\ref{tab:case} gives some cases of different prediction results for a given word sequence. The proceeding context is `我(I) 打(play)' and `商店(store) 不能(can't use) 微信(WeChat)'. And the ground truth is `电话 (phone) 游戏 (games)' and `支付(pay)', respectively. From the results, we can see that statistical language models can only give some frequent result, since they are estimating the probability based on the counting.
For example, the n-gram of `打游戏(play games)' and `微信支付(WeChat Pay)' are rare in the training data, therefore statistical language models will miss these positive words, and can only output some popular positive words such as `电话(phone)' and `群(group)'. While for neural language model, we can see that they can output many more reasonable words, such as `游戏(games)',`疫苗(vaccine)' for `我打(I play(do))', and `密码(password)',`账号(ID)' for `微信(WeChat)'. This is mainly because they can leverage the distributed representations to construct connections between similar words, thus help to expand the candidates.
Therefore, statistical and neural language models both have their own advantages: statistical ones can accurately predict the popular words, while neural ones have the ability to provide more chances to target the rare words to tackle the sparse problem.
Furthermore, the influence of smoothed strategy on the sparse problem is limited, since n-gram-KN provide exactly the same results with n-gram model, only with different order.
\end{CJK*}

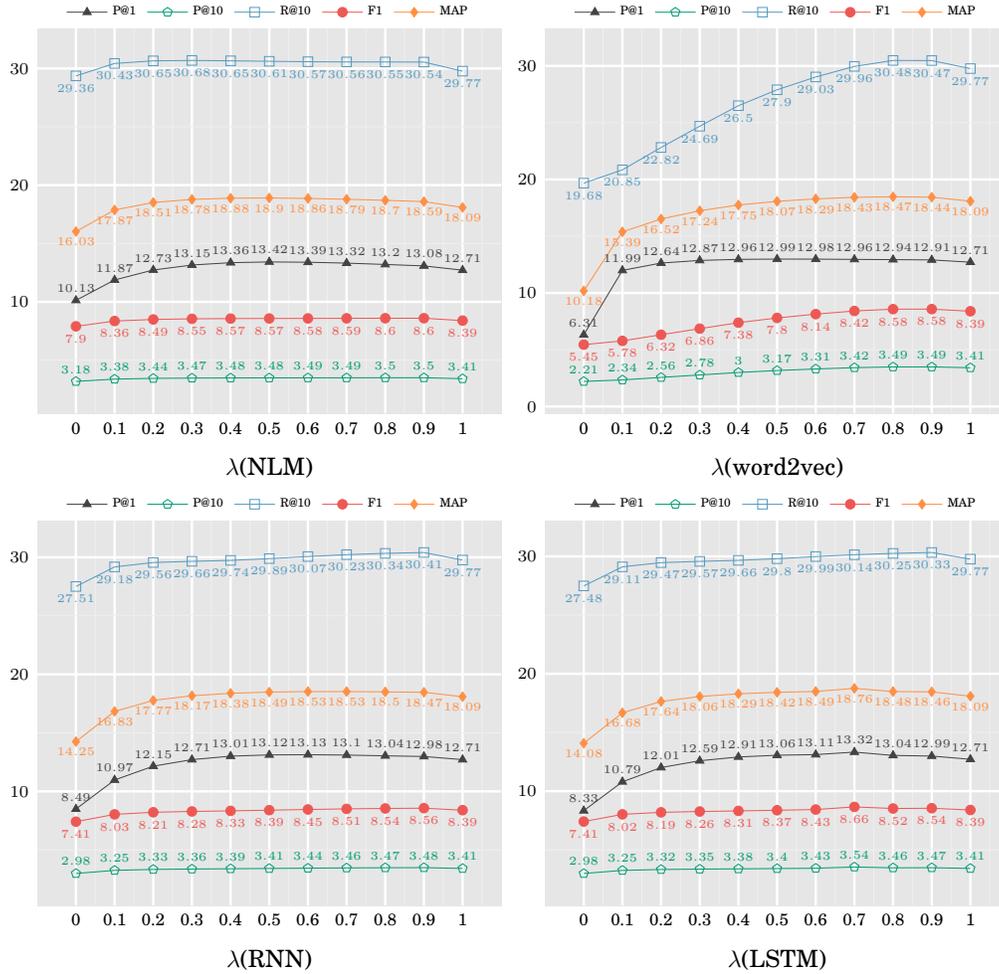
\begin{figure}
\centering
  \begin{tikzpicture}[scale=0.9]
    \begin{axis}[
      xlabel=$\lambda$(NLM),
      xticklabels={0,0.1,0.2,0.3,0.4,0.5,0.6,0.7,0.8,0.9,1},
      xtick={0,1,2,3,4,5,6,7,8,9,10},
      legend style={
          font=\tiny,
          draw=none,
          legend columns=-1,
          at={(0.5,1)},
          anchor=south,
          /tikz/every even column/.append style={column sep=0.9mm}
        },
        ymajorgrids,
        major grid style={draw=white},
        y axis line style={opacity=0},
        tickwidth=0pt,
        nodes near coords,
        every node near coord/.append style={anchor=north, font=\fontsize{4pt}{4pt}\selectfont},
        legend entries = {P@1,P@10,R@10,F1,MAP},
        every tick label/.append style={font=\scriptsize}
        ]
       \addplot[color=tuatara,mark=triangle*, every node near coord/.append style={anchor=south}] coordinates {
(10,12.713300)
(9,13.079200)
(8,13.2019)
(7,13.317700)
(6,13.392800)
(5,13.421000)
(4,13.355100)
(3,13.153900)
(2,12.729200)
(1,11.874300)
(0,10.126200)
      };
      \addplot[color=free_speech_aquamarine, mark=pentagon, every node near coord/.append style={anchor=south}] coordinates {
(10,3.413860)
(9,3.496190)
(8,3.49512)
(7,3.491640)
(6,3.487500)
(5,3.483460)
(4,3.477510)
(3,3.466730)
(2,3.441380)
(1,3.379310)
(0,3.181190)
      };

      \addplot[color=shakespeare,mark=square] coordinates {
(10,29.765900)
(9,30.544900)
(8,30.5521)
(7,30.555300)
(6,30.574000)
(5,30.610700)
(4,30.652400)
(3,30.683000)
(2,30.647800)
(1,30.425200)
(0,29.356800)
      };
      \addplot[color=flamingo, mark=*, every node near coord/.append style={anchor=north}] coordinates {
(10,8.3911)
(9,8.5967)
(8,8.5950)
(7,8.5882)
(6,8.5809)
(5,8.5748)
(4,8.5651)
(3,8.5451)
(2,8.4923)
(1,8.3560)
(0,7.9016)
      };
	\addplot[color=sun_shade, mark=diamond*, every node near coord/.append style={yshift=0.3mm,anchor=north}] coordinates {
(10,18.088500)
(9,18.588300)
(8,18.6972)
(7,18.791800)
(6,18.858900)
(5,18.899300)
(4,18.880600)
(3,18.776700)
(2,18.511700)
(1,17.872700)
(0,16.025400)
      };
    \end{axis}
  \end{tikzpicture}
  \begin{tikzpicture}[scale=0.9]
    \begin{axis}[
      xlabel=$\lambda$(word2vec),
      xticklabels={0,0.1,0.2,0.3,0.4,0.5,0.6,0.7,0.8,0.9,1},
      xtick={0,1,2,3,4,5,6,7,8,9,10},
      legend style={
          font=\tiny,
          draw=none,
          legend columns=-1,
          at={(0.5,1)},
          anchor=south,
          /tikz/every even column/.append style={column sep=0.9mm}
        },
        ymajorgrids,
        major grid style={draw=white},
        y axis line style={opacity=0},
        tickwidth=0pt,
        nodes near coords,
        every node near coord/.append style={anchor=north, font=\fontsize{4pt}{4pt}\selectfont},
        legend entries = {P@1,P@10,R@10,F1,MAP},
        every tick label/.append style={font=\scriptsize}
        ]
       \addplot[color=tuatara,mark=triangle*, every node near coord/.append style={anchor=south}] coordinates {
(10,12.7133)
(9,12.9124)
(8,12.9381)
(7,12.9594)
(6,12.9821)
(5,12.9864)
(4,12.9609)
(3,12.8685)
(2,12.6382)
(1,11.9903)
(0,6.313)
      };
      \addplot[color=free_speech_aquamarine, mark=pentagon, every node near coord/.append style={anchor=south}] coordinates {
(10,3.4139)
(9,3.4888)
(8,3.4874)
(7,3.4208)
(6,3.3060)
(5,3.1678)
(4,2.9964)
(3,2.7801)
(2,2.5620)
(1,2.3429)
(0,2.209)
      };

      \addplot[color=shakespeare,mark=square] coordinates {
(10,29.7659)
(9,30.4747)
(8,30.4833)
(7,29.9595)
(6,29.0284)
(5,27.9003)
(4,26.4988)
(3,24.6914)
(2,22.8196)
(1,20.8464)
(0,19.675)
      };
      \addplot[color=flamingo, mark=*, every node near coord/.append style={anchor=north}] coordinates {
(10,8.3911)
(9,8.5783)
(8,8.5759)
(7,8.4153)
(6,8.1368)
(5,7.8011)
(4,7.3847)
(3,6.8570)
(2,6.3224)
(1,5.7807)
(0,5.4515)
      };
	\addplot[color=sun_shade, mark=diamond*, every node near coord/.append style={yshift=0.3mm,anchor=north}] coordinates {
(10,18.0885)
(9,18.4377)
(8,18.4728)
(7,18.4280)
(6,18.2896)
(5,18.0723)
(4,17.7516)
(3,17.2433)
(2,16.5177)
(1,15.3872)
(0,10.181)
      };
    \end{axis}
  \end{tikzpicture}
  \begin{tikzpicture}[scale=0.9]
    \begin{axis}[
      xlabel=$\lambda$(RNN),
      xticklabels={0,0.1,0.2,0.3,0.4,0.5,0.6,0.7,0.8,0.9,1},
      xtick={0,1,2,3,4,5,6,7,8,9,10},
      legend style={
          font=\tiny,
          draw=none,
          legend columns=-1,
          at={(0.5,1)},
          anchor=south,
          /tikz/every even column/.append style={column sep=0.9mm}
        },
        ymajorgrids,
        major grid style={draw=white},
        y axis line style={opacity=0},
        tickwidth=0pt,
        nodes near coords,
        every node near coord/.append style={anchor=north, font=\fontsize{4pt}{4pt}\selectfont},
        legend entries = {P@1,P@10,R@10,F1,MAP},
        every tick label/.append style={font=\scriptsize}
        ]
       \addplot[color=tuatara,mark=triangle*, every node near coord/.append style={anchor=south}] coordinates {
(10,12.71330)
(9,12.98110)
(8,13.03530)
(7,13.10240)
(6,13.13060)
(5,13.11530)
(4,13.00870)
(3,12.71130)
(2,12.15390)
(1,10.97230)
(0,8.49055)
      };
      \addplot[color=free_speech_aquamarine, mark=pentagon, every node near coord/.append style={anchor=south}] coordinates {
(10,3.41386)
(9,3.48281)
(8,3.47351)
(7,3.45875)
(6,3.43702)
(5,3.41168)
(4,3.38585)
(3,3.36348)
(2,3.33032)
(1,3.25120)
(0,2.98432)
      };

      \addplot[color=shakespeare,mark=square] coordinates {
(10,29.76590)
(9,30.41160)
(8,30.33800)
(7,30.22750)
(6,30.06610)
(5,29.88890)
(4,29.73610)
(3,29.65820)
(2,29.55620)
(1,29.18050)
(0,27.50680)
      };
      \addplot[color=flamingo, mark=*, every node near coord/.append style={anchor=north}] coordinates {
(10,8.3911)
(9,8.5630)
(8,8.5405)
(7,8.5052)
(6,8.4533)
(5,8.3933)
(4,8.3336)
(3,8.2848)
(2,8.2130)
(1,8.0346)
(0,7.4109)
      };
	\addplot[color=sun_shade, mark=diamond*, every node near coord/.append style={yshift=0.3mm,anchor=north}] coordinates {
(10,18.08850)
(9,18.46840)
(8,18.49920)
(7,18.52880)
(6,18.52730)
(5,18.48500)
(4,18.38100)
(3,18.17160)
(2,17.77140)
(1,16.83220)
(0,14.25160)
      };
    \end{axis}
  \end{tikzpicture}
  \begin{tikzpicture}[scale=0.9]
    \begin{axis}[
      xlabel=$\lambda$(LSTM),
      xticklabels={0,0.1,0.2,0.3,0.4,0.5,0.6,0.7,0.8,0.9,1},
      xtick={0,1,2,3,4,5,6,7,8,9,10},
      legend style={
          font=\tiny,
          draw=none,
          legend columns=-1,
          at={(0.5,1)},
          anchor=south,
          /tikz/every even column/.append style={column sep=0.9mm}
        },
        ymajorgrids,
        major grid style={draw=white},
        y axis line style={opacity=0},
        tickwidth=0pt,
        nodes near coords,
        every node near coord/.append style={anchor=north, font=\fontsize{4pt}{4pt}\selectfont},
        legend entries = {P@1,P@10,R@10,F1,MAP},
        every tick label/.append style={font=\scriptsize}
        ]
       \addplot[color=tuatara,mark=triangle*, every node near coord/.append style={anchor=south}] coordinates {
(10,12.71330)
(9,12.98610)
(8,13.04210)
(7,13.32270)
(6,13.11170)
(5,13.05620)
(4,12.91060)
(3,12.58620)
(2,12.00780)
(1,10.79470)
(0,8.33090)
      };
      \addplot[color=free_speech_aquamarine, mark=pentagon, every node near coord/.append style={anchor=south}] coordinates {
(10,3.41386)
(9,3.47428)
(8,3.46450)
(7,3.53677)
(6,3.42819)
(5,3.40194)
(4,3.37782)
(3,3.35489)
(2,3.32224)
(1,3.24573)
(0,2.98383)
      };

      \addplot[color=shakespeare,mark=square] coordinates {
(10,29.76590)
(9,30.33260)
(8,30.25470)
(7,30.13990)
(6,29.98660)
(5,29.80220)
(4,29.66150)
(3,29.57480)
(2,29.46850)
(1,29.11140)
(0,27.48450)
      };
      \addplot[color=flamingo, mark=*, every node near coord/.append style={anchor=north}] coordinates {
(10,8.3911)
(9,8.5417)
(8,8.5181)
(7,8.6554)
(6,8.4314)
(5,8.3692)
(4,8.3136)
(3,8.2632)
(2,8.1922)
(1,8.0201)
(0,7.4089)
      };
	\addplot[color=sun_shade, mark=diamond*, every node near coord/.append style={yshift=0.3mm,anchor=north}] coordinates {
(10,18.08850)
(9,18.45730)
(8,18.48400)
(7,18.75750)
(6,18.49130)
(5,18.41900)
(4,18.29260)
(3,18.06270)
(2,17.63810)
(1,16.67500)
(0,14.07960)
      };
    \end{axis}
  \end{tikzpicture}
  \caption{\label{fig:lambda} Influences of different $\lambda$ in combining n-gram and neural model.}
\end{figure}
\subsection{Combination of Statistical and Neural Language Models}
According to the above analysis that statistical and neural language models usually give different results, we give a hybrid model as follows.
\begin{eqnarray*}
P(w_t|C){=}\lambda P_{n}(w_t|C){+}(1{-}\lambda)P_{d}(w_t|C),
\end{eqnarray*}
where $P_{n}$ and $P_d$ stands for the conditional probability produced by statistical and neural language models, respectively. $\lambda$ is a tradeoff factor in the range of [0,1].
When $\lambda=1$, it reduces to statistical language model. While if $\lambda=0$, it reduces to neural language model.

We first study the influences of different $\lambda$ ranging from 0 to 1 with step $0.1$, on the validation set of CIR to see whether the combination will help for the Chinese input recommendation task. Figure~\ref{fig:lambda} show the results of different $\lambda$ when combing n-gram with NLM, word2vec, RNN and LSTM respectively. From the results, we can see that the performances are changing in a similar trend, i.e. first increase and then drop. The best $\lambda$ for combing n-gram and NLM is $0.5$, the best one for combing n-gram and word2vec is $0.8$, the best $\lambda$ for combing n-gram and RNN is $0.9$, while the best one for combing n-gram and LSTM is $0.9$. Therefore, we can see that the hybrid model can indeed obtain better results. These parameters are used for further comparison.

Table~\ref{tb:combination} gives the comparison results between the combination approach with the single ones. We can see that the combination approach significantly improve the results. Taking MAP as an example, the improvement of the best combination method (i.e.~NLM+n-gram/18.78) over the best statistical method (i.e.~n-gram/18.09) is 3.8\%. While the improvement over the best neural method (i.e.~NLM/16.03) is 17.2\%. This is accordant with the experimental findings that neural language models can provide suitable similar words for candidate to tackle the sparse problem, as shown in the above case studies.

Considering the above results that word2vec is different from other neural language models, and the overlapping is very small, we propose to combine n-gram, word2vec, and other neural language models. Therefore, we can obtain the following hybrid model:
\begin{eqnarray*}
P(w_t|C){=}\lambda_1 P_{n}(w_t|C){+}\lambda_2P_{w}(w_t|C)+(1-\lambda_1-\lambda_2)P_{d}(w_t|C),
\end{eqnarray*}
where $P_{n}$, $P_w$ and $P_d$ stands for the conditional probability produced by n-gram ,word2vec and other neural language models, respectively. $\lambda_1,\lambda_2$ are tradeoff parameters in the range of [0,1]. In our experiments, we also tune these parameters in the validation set, and only report the best results with $\lambda_1=0.3$ and $\lambda_2=0.2$. The experimental results are shown in Table \ref{tb:combination}. We can see that the performance is improved significantly. Taking MAP as an example, the improvement(i.e.~NLM+n-gram+word2vec/19.59) over the best statistical method(i.e.~n-gram/18.09) is  8.3\%.

\subsection{Discussions on Different Neural Language Models}
Furthermore, we conduct a discussion on different neural language models to provide some more insights. Since word2vec is proposed as a simplified version of NLM, and RNN can be viewed as more complicated than NLM, we conduct the discussions on word2vec vs. NLM, and RNN vs. NLM, respectively.

\subsubsection{Word2Vec vs. NLM}
The experiments show that the combination of word2vec+n-gram works worse than that of NLM+n-gram. We analyzed the data and found that the failure of word2vec+n-gram may be caused by the fact that word2vec did not consider the orders of the given context, since they are usually using an average operation to obtain the context vector in word2vec. However, such order information is usually crucial for the task of language modeling. To reflecting such order information while keeping the efficiency advantage of word2vec, we propose to modify the original word2vec to a weighted version. Specifically, the new feature vector of each context word $v^c(w_j),j=t-L,\cdots,t-1$ is defined as the the following form:
\begin{displaymath}
v^c(w_j)=\frac{2(t-j)\cdot v^c(w_j)}{L(1+L)},
\end{displaymath}
where $v^c(w_j)$ stands for the feature vector of $w_j$ in original word2vec.
The experimental results in Table~\ref{tb:combination} show that the combination of weighted word2vec and n-gram can perform better than that of word2vec and n-gram.

\begin{CJK*}{UTF8}{gbsn}
Specifically, we give a specific case for further explanation. Given a word sequence `你(you)没有(not have it)就(is)-行(okay)', if we adopt weighted word2vec, the predicted words will be `行了(okay)可以了(alright)行(okay)可以(alright)好了(good)'. Compared with the results produced by word2vec `这样(doing this)那么(that)必要(need)发言权(right to speak)这么(so)', we can see that the words related to the latter word `就(is)' has been ranked on top positions. Therefore, the order information of contexts have been taken into account in the model.
\end{CJK*}

\subsubsection{NLM vs. RNN}
The experiments show that RNN+ngram is worse than NLM+ngram. We give some explanations as follows. RNN compresses the context information into a hidden layer, while NLM has a fully connected with the context. Therefore the advantage of RNN lies in the modeling for long context, since it can capture the long term dependencies. But in our data, each sentence has 4.9 words on average and NLM has a window size 6 in our experiments. That is to say, NLM has ability to control most situations(referred to Table\ref{tb:Sdistribution}).

\section{Conclusions and Future Work}\label{sec:conclusion}
In this paper, we conduct an empirical study on a large scale real data of Chinese input recommendation task. The experimental findings show that: (1) Statistical and neural language models have their own advantages, i.e.~statistical ones can provide more accurate results by utilizing counting as the estimation foundation, while neural ones can alleviate the sparsity problem by providing more similar results; (2) The combination of the two approach will improve the results of Chinese input recommendation, since the overlap between them are relatively small;
Word2vec is different from NLM and RNN due to the negative sampling strategy, and the combination of n-gram, word2vec and other neural language models will further improve the results.

For the future work, we will further investigate the different neural language models, such as whether word2vec provides some diverse results, which may be helpful for tail users. We will also study the efficiency issue of different neural language models, which is more important for the application on mobile.
\bibliographystyle{ACM-Reference-Format-Journals}
\bibliography{tois2016}

\end{document}